# Transformer-Based Self-Supervised Learning for Histopathological Classification of Ischemic Stroke Clot Origin

Kun-Hao Yeh, MS [*]; Mohamed Sobhi Jabal, MD, MS [1]; Vikash Gupta, PhD [2]; David F. Kallmes, MD [1]; Waleed Brinjikji, MD [1]; Barbaros Selnur Erdal, DDS, PhD [2]

**Affiliations:**

[1]Department of Radiology, Mayo Clinic, Rochester, MN, USA

[2]Center for Augmented Intelligence in Imaging, Mayo Clinic, Florida, MN, USA

[*]Independent Scholar

**Corresponding author:**
Mohamed Sobhi Jabal
Department of Radiology
Mayo Clinic, Rochester, MN
Email: m.sobhi.jabal.research@gmail.com

# Abstract

**Background and Purpose**: Identifying the thromboembolism source in ischemic stroke is crucial for treatment and secondary prevention yet is often undetermined. This study describes a self-supervised deep learning approach in digital pathology of emboli for classifying ischemic stroke clot origin from histopathological images.

**Methods**: The dataset included whole slide images (WSI) from the STRIP AI Kaggle challenge, consisting of retrieved clots from ischemic stroke patients following mechanical thrombectomy. Transformer-based deep learning models were developed using transfer learning and self-supervised pretraining for classifying WSI. Customizations included an attention pooling layer, weighted loss function, and threshold optimization. Various model architectures were tested and compared, and model performances were primarily evaluated using weighted logarithmic loss.

**Results**: The model achieved a logloss score of 0.662 in cross-validation and 0.659 on the test set. Different model backbones were compared, with the swin_large_patch4_window12_384 showed higher performance. Thresholding techniques for clot origin classification were employed to balance false positives and negatives.

**Conclusion**: The study demonstrates the extent of efficacy of transformer-based deep learning models in identifying ischemic stroke clot origins from histopathological images and emphasizes the need for refined modeling techniques specifically adapted to thrombi WSI. Further research is needed to improve model performance, interpretability, validate its effectiveness. Future enhancement could include integrating larger patient cohorts, advanced preprocessing strategies, and exploring ensemble multimodal methods for enhanced diagnostic accuracy.

# Introduction

The global burden of ischemic stroke is one the most significant and poses major mortality and morbidity rates that affect healthcare systems worldwide (1,2). One of the fundamental challenges in managing this disease is identifying the thrombus source, a critical aspect influencing both treatment decisions and secondary prevention strategies (3).

Stroke-inducing emboli primarily originate from areas such as the left atrial appendage or ventricle or atherosclerosis of the internal carotid artery (4). Nevertheless, for a notable portion of patients, the embolic source remains undetermined, leading to a diagnosis of embolic stroke of unknown source (ESUS) (5). Such patients are often discharged with a level of uncertainty regarding the optimal secondary prevention management plan.

The diagnostic workup of stroke etiology remains an indispensable cornerstone in decreasing the mortality and morbidity of patients with ESUS (7). Advancements in diagnostic technologies using cerebrovascular imaging via CT or MRI angiography, echocardiography, and ECG monitoring have revolutionized our understanding of underlying stroke causes and have facilitated the creation of more effective therapeutic interventions (8,9).

Innovative research strategies involving the histological examination of emboli obtained from thrombectomies offer promising insights (10). This is especially relevant for patients with large vessel occlusion, where an embolic origin for the stroke is highly probable. However, given the recent popularity of this approach, the optimal methodology for histopathological examination of the clot specimen is well established (11).

The robustness of deep learning (DL) techniques applied across various domains from image classification has shown considerable promise in the medical imaging field (12,13), with an increasing body of evidence illustrating the potential of deep learning in enhancing diagnostic accuracy and augmenting human interpretation of histological image through digital pathology processing (14,15).

To address this stroke etiology challenge, the " Mayo Clinic - STRIP AI" prized Kaggle competition was organized for the classification of ischemic stroke clot origin using machine learning (ML) applied to digital pathology whole slide images (16). The competition brought together ML expertise through crowdsourcing to develop and evaluate DL models to classify clot origin (17). In this study, we detailed the development of the prized solution, in which deep learning was deployed to analyze the digital pathology images of stroke thromboemboli for their origin classification, with an extended analysis to include additional insights and evaluation of the methodology and metrics related to the solution (18,19).

## Materials and Methods

*Dataset*

The dataset incorporated whole slide images obtained through the Stroke Thromboembolism Registry of Imaging and Pathology – Artificial Intelligence (STRIP AI) Kaggle challenge. STRIP is a multi-institutional biobank of clots from ischemic large vessel occlusion stroke patients admitted between 2016 and 2020, with their corresponding digital pathology images. Registry participants are adults (18 or older) who received mechanical thrombectomy for acute ischemic stroke management and had clot material extracted during the procedure. The procedure's technique varied, being tailored to each operator's preference.

Thrombus processing and histological analysis were carried out in as follows: after formalin fixation in the home institution, the bio samples were then sent to a central lab for standard processing and paraffin embedding. The clots were scanned, assessed, and stained by histopathologists.

*Data Preprocessing*

Given the high resolution of WSI, the slides were divided into tiles with the dimensions of (1024 x 1024) each and three-color channels: blue, red, and green. The top darkest tiles were selected from each slide.

*Model Development*

With regard to the modeling process, a combination of deep supervised and self-supervised learning was adopted. A transfer learning-based method was used as the backbone of the model using swin_large_patch4_window12_384 [--], which represents a hierarchical vision transformer using shifted windows applied to computer vision.

The reason for this model choice as a development model is its good performance on Imagenet reaching 87% accuracy, and the GPU's 700 MB, which makes it implementable on Kaggle kernels.

Based on this model architecture, further finetuning was done with regard to the histology classification task producing probabilities of cardioembolic and large artery atherosclerosis for each slide.

For each WSI, the 16 tiles were fed as single input to the model for the clot origin prediction.

Loss-weighted multi-class logarithmic loss was used for model training and validation.

$$\text{Log Loss} = -\left(\frac{\sum_{i=1}^{M} w_i \cdot \sum_{j=1}^{N_i} \frac{y_{ij}}{N_i} \cdot \ln p_{ij}}{\sum_{i=1}^{M} w_i}\right)$$

5-fold stratified grouped K-Fold was also implemented, which was stratified by class and grouped by patient, as some patients had multiple slides pertaining to them.

The classification head was customized to adapt and optimize for the task in question. Self-supervised pretraining was applied using mocov3.

The initial model batch size was 16 x 3 x 384 (h) x 384 (w) for the input, with an output batch size of 16 x L x 1536 (embed size). Instead of implementing average pooling of all the sub-tiles and all the batches, it was replaced by attention pooling with learnable weights to assign different weights to different tiles and locations within each tile. This gives the model the freedom to have different weighted sums for the embedded vectors of the batches.

Through a final fully connected layer at the end, the probabilities of the two clot origin classes were predicted, which were then used to calculate the logarithmic loss.

Prior to model finetuning for this classification task, the initial customized Imagenet pretrained model was further pretrained on this histology task using self-supervised learning with mocov3 which takes random crops of the images, perform significant data augmentation process in relation to brightness, contrast, random masking, blurring, cutout, rotation, and shift. The model then learns to identify whether two crops belong to the same image. This contrastive learning method helps the model capture discriminating features in this data domain. Overall model training time is around 2 to 3 hours.

## Results

The transformer-based model swin_large_patch4_window12_384 achieved a logloss score of 0.662 in the cross-validation and logloss of 0.659 on the test set, topping private leaderboard, showcasing the improvement and the effectiveness of the concept.

Additional experimental results included the following: Initial modeling with 5-fold resulted in cv logloss mean=0.69, std=0.04. Model with 5-fold average and customized classification head: cv mean=0.66, std=0.03. Model 5-fold average with customized classification head and mocov3 pretraining: cv mean=0.66, std=0.015 (more stable across folds). Performance scores of other backbones with the same customized classification head and mocov3 pretraining include vit256: 0.683 ±0.03. coat_lite_medium: 0.673 ±0.04. coatnet_rmlp_1_rw_224: 0.687 ±0.03. convnext_large_384_in22ft1k: 0.674 ±0.03. All model backbones are based on 'timm' computer vision model implementation and are all pretrained from ImageNet with further in-domain pretraining and finetuning on this task.

Weighted F1, precision, and recall curves are provided in the graph shown in Figure 2A, based on different thresholding for CE origin prediction. The best threshold is 0.4 to achieve the best weighted f1 score: 0.6781. At 0.4, in case of prediction probability for CE being > 0.4, the prediction is marked as CE; otherwise, LAA would be inferred. This allows for detailing true and false rates of positive and negative predictions; then, we use them according to the number of samples to get weighted F1, precision, and recall. With optimal thresholding, the confusion matrix for detailed performance is shown in Figure 2B.

## Discussion

The application of transformer-based deep learning models for the histopathological classification of ischemic stroke clot origin demonstrates promising potential toward addressing the intricate challenge of stroke etiology. Our results underscore the potential of leveraging self-supervised learning paradigms in conjunction with domain-specific data to augment the capabilities of existing diagnostic methodologies. The findings, and more specifically resulting model performance metrics, suggest that when the model predicts a large artery atherosclerotic origin, the ground truth that the clot is indeed of type LAA was very likely, on the other hand, when the predictions pointed towards CE as the source, the likelihood that the origin of the clot would be of type LAA is around 50%.

The demonstrated improvement in logloss scores through the customizations and in-domain pretraining (from a cv mean logloss of 0.69 to 0.66 with a reduced standard deviation of 0.015) accentuates the value of domain adaptation and the potential for further enhancements in model performance with more refined domain-centric pretraining strategies. The comparative analysis of different model backbones provides valuable insight into the tradeoffs between different architectures and their effectiveness in this specific task. Among the models evaluated, the swin_large_patch4_window12_384 backbone demonstrated superior performance, potentially due to its hierarchical representation learning capability which may capture the complex underlying patterns in histopathological images more effectively. The choice of loss function and evaluation strategy are crucial aspects that warrant further exploration to ensure the reliability and interpretability of the model's predictions in clinical settings. Moreover, the thresholding technique employed for classifying clot origins as Cardioembolic (CE) or Large Artery Atherosclerosis (LAA) evidenced a pragmatic approach to balancing the tradeoff between false positives and false negatives, which is paramount in clinical decision-making scenarios where misclassifications could have significant implications on patient management and outcomes.

The relatively quick training time of around 2 to 3 hours and the manageable GPU memory requirement (700 MB) of the model facilitate its deployment in real-world, resource-constrained settings, opening up avenues for further on-site evaluations and potentially real-time applications in aiding histopathological examinations.

While the applied trained models were slightly predictive of the stroke clot origin in the classification task and achieved reasonably fair however non-optimal performance, several potential avenues could be considered for future improvement, such as acquiring additional WSI data from more patients would likely enhance model performance, utilizing batch-level detailed segmentation of blood cells and clot components with pretrained models could provide extra informative input channels, transfer learning from models pretrained on histology and blood cell WSIs, rather than ImageNet, may boost results, and multitask learning approaches that jointly predict clot source along with other relevant labels could allow the model to learn useful features and representations. Overall, there remains ample opportunity to refine and advance deep learning approaches with additional data and optimized modeling strategies. We expect further gains in predictive accuracy of determining ischemic stroke clot origin which remains an unresolved clinically significant task.

Modeling time and resource expenses present another challenge when dealing with large WSI model training. There exists a tradeoff between the model size and inference time, as smaller models might have reduced performance on Imagenet and lack the ability to learn complex patterns possibly resulting in overfitting after finetuning for our classification task. The data-preprocessing part was the costliest in terms of time and compute resources and could potentially be sped up using multi-threading and multiprocessing, in addition to using lower-level compiled languages and more graphical and computing hardware infrastructure.

Despite promising initial results, several aspects warrant further investigation. For instance, the scalability and effectiveness of the model on larger and more diverse datasets in different clinical settings remain to be explored. Additionally, the interpretability of the predictions is crucial for clinical adoption and needs to be addressed. Future work could also explore ensemble methods, combining the strengths of different architectures, or integrating multimodal data sources to enhance further the diagnostic accuracy and robustness of the proposed approach.

**Conclusion**

This study showcases the promising application of deep learning, specifically transformer-based models and self-supervised learning, in identifying ischemic stroke clot origins from histopathological images. Despite achieving improved logloss scores and reasonable predictive performance, the study highlights areas for improvements, including the need for more comprehensive data and refined modeling techniques. The feasibility of clinical deployment is suggested by the model's quick training time and low GPU requirements. Broader validation across varied clinical settings combined with efforts to enhance model interpretability could accelerate the translation of such assisted pathology solutions into routine clinical practice, delivering data-driven insights to guide targeted therapeutic interventions for improved patient outcomes. Recommended future directions include integrating additional patient data, employing advanced pretraining strategies, and exploring multitask learning and ensemble methods.

**Table 1**. Results of different experimental methods implemented on the public dataset.

| Model No | Model Backbone | Experiment Setting | logloss (CV) |
|---|---|---|---|
| 1 | swin_large_patch4_window12_384 | Default Setting | 0.69 (0.04) |
| 2 | swin_large_patch4_window12_384 | Customized Classification Head | 0.66 (0.03) |
| 3 | swin_large_patch4_window12_384 | Customized Classification Head + In-Domain Mocov3 Pretraining | 0.66 (0.015) |
| 4 | vit256 | Customized Classification Head + In-Domain Mocov3 Pretraining | 0.683 (0.03) |
| 5 | coat_lite_medium | Customized Classification Head + In-Domain Mocov3 Pretraining | 0.673 (0.04) |
| 6 | coatnet_rmlp_1_rw_224 | Customized Classification Head + In-Domain Mocov3 Pretraining | 0.687 (0.03) |
| 7 | convnext_large_384_in22ft1k | Customized Classification Head + In-Domain Mocov3 Pretraining | 0.674 (0.03) |

A

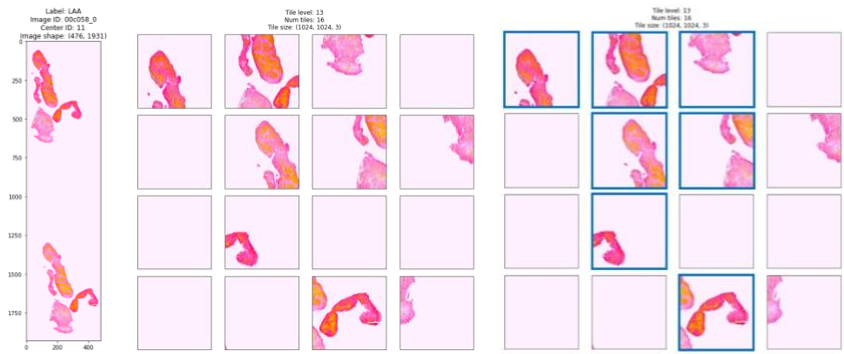

B

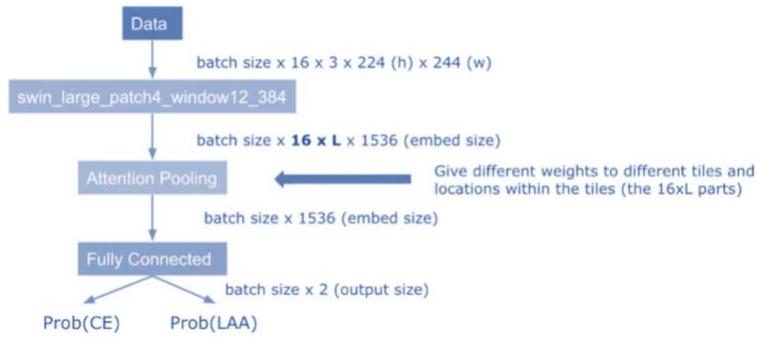

**Figure 1.** WSI tiling, tile selection process (A), and modeling steps (B) for binary clot source classification.

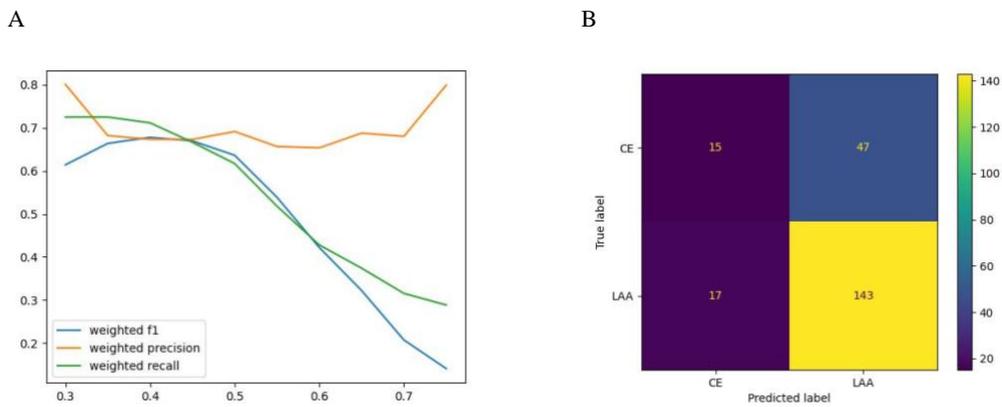

**Figure 2.** Evaluation metrics combining weighted F1 score, precision, and recall curves (A) and confusion matrix of the final classification metrics (B).